\documentclass[conference]{IEEEtran}
\IEEEoverridecommandlockouts
\usepackage{cite}
\usepackage{amsmath,amssymb,amsfonts}
\usepackage{algorithmic}
\usepackage{graphicx}
\usepackage{textcomp}
\usepackage{xspace}
\usepackage{amsmath}
\usepackage{enumitem}
\usepackage{xcolor}
\usepackage{booktabs}
\usepackage[short]{optidef}
\usepackage{caption}
\usepackage{multirow}
\usepackage{subcaption}
\usepackage{bm}
\def\BibTeX{{\rm B\kern-.05em{\sc i\kern-.025em b}\kern-.08em
    T\kern-.1667em\lower.7ex\hbox{E}\kern-.125emX}}

\def\BibTeX{{\rm B\kern-.05em{\sc i\kern-.025em b}\kern-.08em
    T\kern-.1667em\lower.7ex\hbox{E}\kern-.125emX}}

\newcommand{\method}{\texttt{Multi-pofo}\xspace }

\def\BibTeX{{\rm B\kern-.05em{\sc i\kern-.025em b}\kern-.08em
    T\kern-.1667em\lower.7ex\hbox{E}\kern-.125emX}}

\def\BibTeX{{\rm B\kern-.05em{\sc i\kern-.025em b}\kern-.08em
    T\kern-.1667em\lower.7ex\hbox{E}\kern-.125emX}}

\begin{document}


\title{
A Unified Energy Management Framework for Multi-Timescale Forecasting in Smart Grids
}

\author{\IEEEauthorblockN{Dafang Zhao$^{\spadesuit,\dagger}$, Xihao Piao$^{\spadesuit}$, Zheng Chen$^{\spadesuit}$, Lingwei Zhu$^{\clubsuit}$, Zhengmao Li$^{\diamondsuit}$, Ittetsu Taniguchi$^{\spadesuit}$}
  \IEEEauthorblockA{$\spadesuit$Osaka University, Japan\\
  $\clubsuit$The University of Tokyo, Japan
  \\
   $\diamondsuit$Aalto University, Finland}\\
\thanks{
$\dagger$ Corresponding author: Dafang Zhao (zhao.dafang@ist.osaka-u.ac.jp)}
}
\maketitle

\begin{abstract}
    Accurate forecasting of the electrical load, such as the magnitude and the timing of peak power, is crucial to successful power system management and implementation of smart grid strategies like demand response and peak shaving.
    In multi-time-scale optimization scheduling, rolling optimization is a common solution.
    However, rolling optimization needs to consider the coupling of different optimization objectives across time scales.
    It is challenging to accurately capture the mid- and long-term dependencies in time series data.
    This paper proposes \method, a multi-scale power load forecasting framework, that captures such dependency via a novel architecture equipped with a temporal positional encoding layer. 
    To validate the effectiveness of the proposed model, we conduct experiments on real-world electricity load data.
    The experimental results show that our approach outperforms compared to several strong baseline methods.
\end{abstract}
    

\section{Introduction}\label{sec:intro}

In recent years, the increasing penetration of renewable energy sources has brought challenges to grid stability.
Load forecasting plays a crucial role in power system operation and planning, as it provides the basis for decision-making, optimal scheduling, and maintenance for the power system~\cite{fang2024improving}. 
Accurate forecasting of electricity demand, magnitude, timing of  peak load  is essential for efficient power system operation and strategic planning.
Major consumption of electricity by e.g. buildings, as well as smart grid technologies like the demand response programs and peak shaving, all require accurate forecasting to properly manage and plan system stability, cost reduction and grid resilience~\cite{Zhao2023}. 

\begin{figure}
    \centering
    \includegraphics[width=\linewidth]{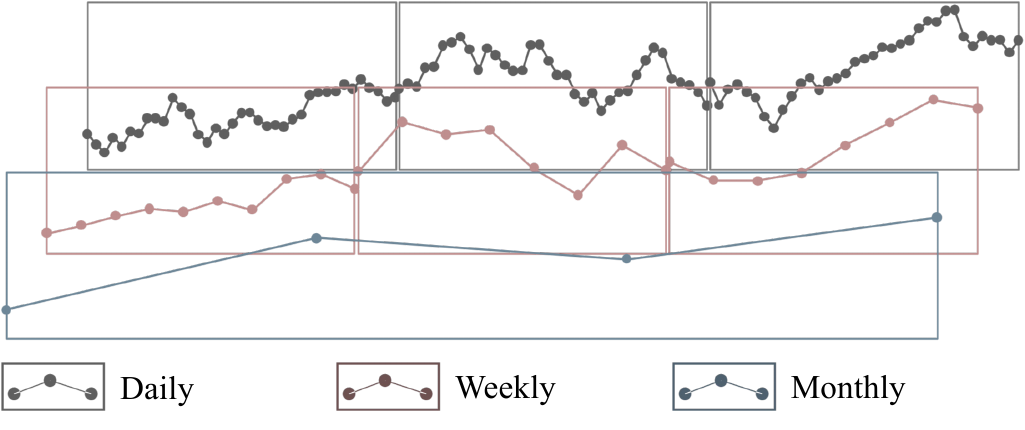}
    \caption{Building load as a superposition of different time scales, such as daily, weekly, monthly trends.}
    \label{fig:mts}
\end{figure}

In practice, multi-timescale forecasting is required to make timely decisions and  support diverse operational needs, from immediate load balancing to long-term capacity planning. 
Building load can be regarded as superposition of linear or nonlinear time series signals of different time scales, ranging from daily, weekly, seasonal patterns to noise~\cite{liMultiScaleDilatedConvolution2024,tan2022multi}, as depicted in Figure~\ref{fig:mts}.
These features often contain complex, nonlinear interactions between short-term fluctuations and long-term trends, making it challenging to build models that can effectively balance between high and low-frequency components. Additionally, the extraction of meaningful patterns across scales demands advanced feature extraction techniques to separate noise from valuable signals. 
As real-world data typically display heterogeneity and non-stationarity, an additional layer of difficulty is imposed on the task and requires computationally efficient models that are both adaptable to changing conditions and robust to noises, which is vital for any downstream tasks.

\begin{figure*}[!t]
    \centering
    \begin{subfigure}[b]{1.1\columnwidth}
         \includegraphics[width=\linewidth]{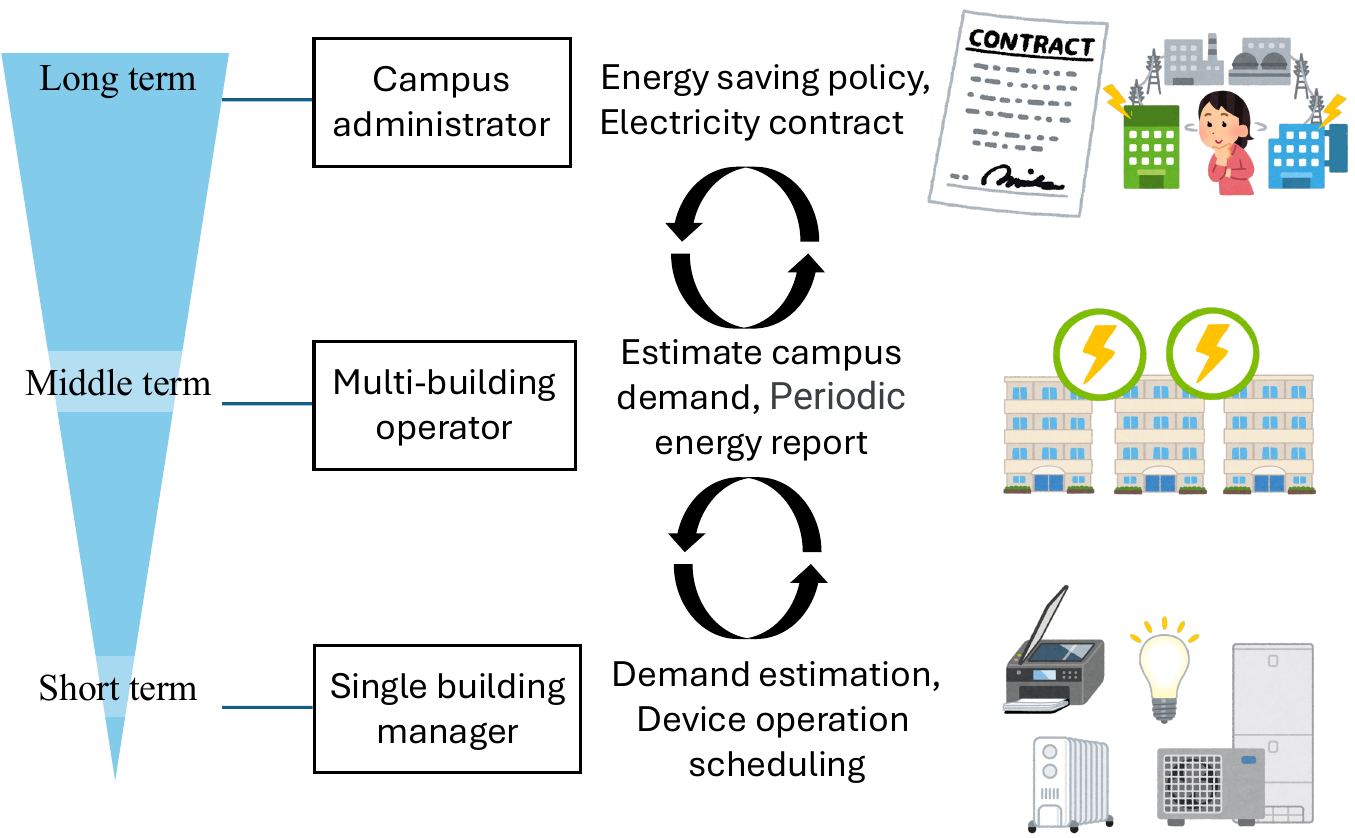}
         \caption{}
         \label{fig:mtst}
    \end{subfigure}
    \hspace{0.3cm}
    \begin{subfigure}[b]{0.88\columnwidth}
        \includegraphics[width=\linewidth]{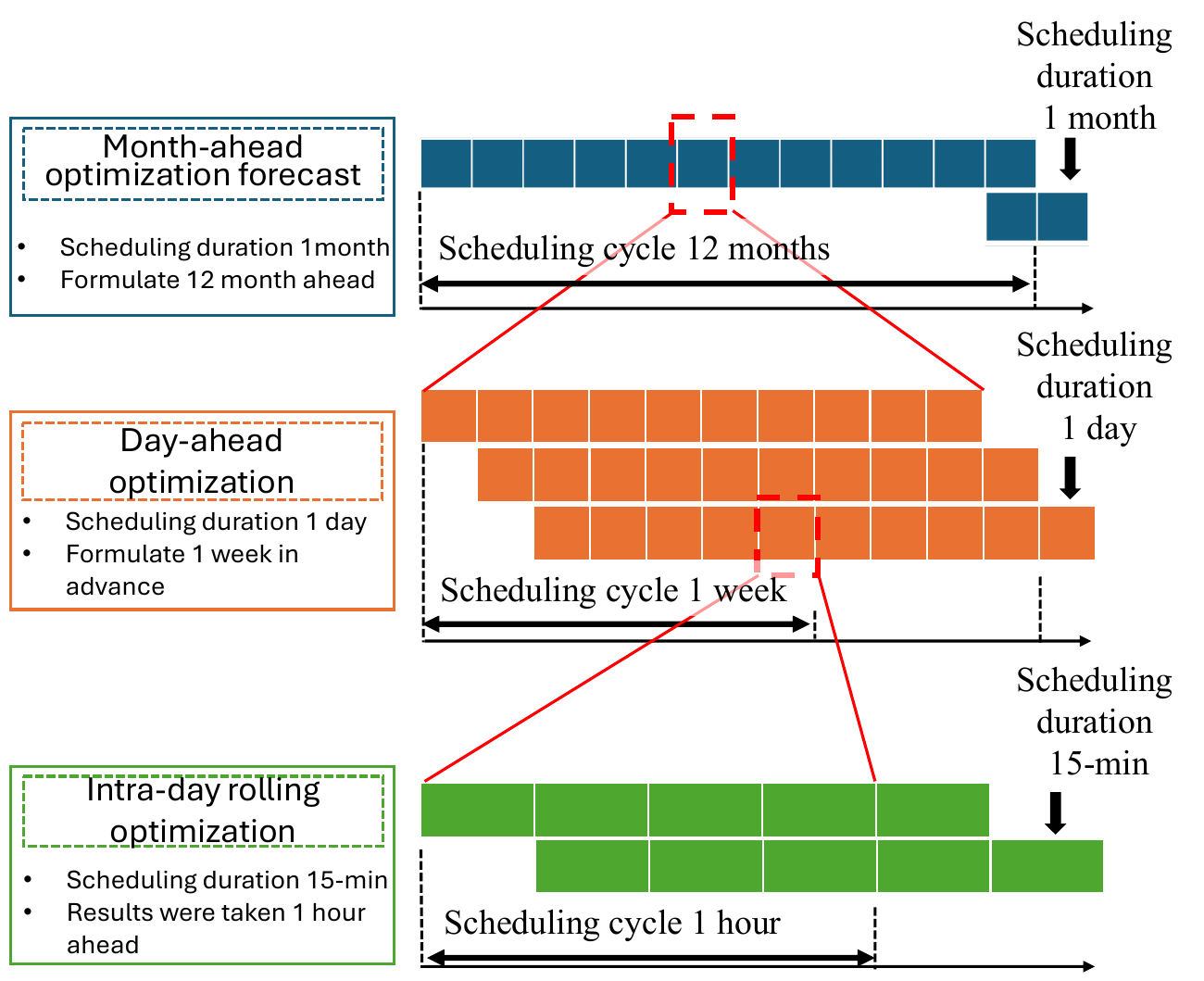}
        \caption{}
        \label{fig:mtso}
    \end{subfigure}
    \caption{Example of (a) campus level multi-timescale energy optimization architecture; (b) multi-timescale energy optimization process.}
\end{figure*}

Recently, there is a growing interest in leveraging deep learning models for enhancing performance in time series forecasting, often with models like the Long-Short Term  Memory (LSTM), Convolutional Neural Networks (CNN), among others~\cite{du2020power,dengMultiScaleConvolutionalNeural2019,louMTSLSTDMMultiTimeScaleLong2022,wu2021autoformer}. 
These models excel at learning nonlinear, robust and general representations that have been proven crucial to capturing diverse patterns within time series data. 
However, a key limitation of current deep learning models in time series forecasting is that they generally require input data to be on a uniform scale, but existing learning frameworks are largely designed for one-to-one mappings~\cite{piaoFredformerFrequencyDebiased2024a}. 
Due to the fixed size and scale of model inputs, they cannot directly accommodate multi-scale data inputs or produce multi-scale outputs \cite{piao2024frednormer}. 

The restriction to a single time scale hinders the ability to capture dependencies across different time scales, particularly when short-, mid-, and long-term relationships coexist in the data.
Moreover, it complicates the optimization scheduling process, as it demands rolling optimization approaches that  account for overlapping objectives across forecast horizons.
Accurately capturing these dependencies across time scales is therefore critical for improving forecast quality and optimizing energy management across multiple temporal resolutions.

To address these challenges, in this paper, we propose \method, a unified framework for energy management that can receive varied time series inputs while providing \textbf{Multi}-scale \textbf{po}wer \textbf{fo}recasting results.
Specifically, \method is a machine learning-based model consisting of three parts.
The first part is a multi-scale embedding that learns scale-specific features while maintaining a unified input format;
The second part consists of a multi-scale encoder that captures common features among multiple scales by sharing the encoder across all scales;
and lastly is a separate prediction module for forecasting.
In summary, our contributions are:
\begin{enumerate}
    \item a scalable framework: \method is designed to seamlessly handle multi-scale and irregular time-series inputs while producing outputs across different scales. This scalability ensures it can adapt to diverse energy management scenarios, from short-term operational adjustments to long-term strategic planning;
    \item a simple yet powerful architecture. By leveraging a simple multi-layer perceptron (MLP) model, \method achieves state-of-the-art forecasting performance with reduced computational complexity.
    We verify that \method outperforms popular deep learning baselines such as CNN-LSTM and BiLSTM~\cite{dengMultiScaleConvolutionalNeural2019,du2020power}.
The simplicity and superior performance of our method render it a suitable choice for real-world deployment.
\end{enumerate}


\section{multi-timescale Energy management}\label{sec:background}

Efficient optimization strategies are crucial for improving energy efficiency, reducing costs, and ensuring grid stability.
Most research primarily focused on formulating efficient day-ahead scheduling plans for energy systems. 
However, the diversity of multi-energy and load characteristics means that relying solely on day-ahead scheduling plans might not always be able to meet optimization objectives.

To address this challenge, multi-timescale optimization strategies have been proposed to optimize energy management across different time scales~\cite{maMultitimescaleOptimizationScheduling2023,wangMultitimeScaleEnergy2024,zhuResearchMultitimeScale2024}. For example, in a campus-level energy management system, the optimization process spans multiple time scales, such as short-, middle- and long-term scheduling, as illustrated in Figure~\ref{fig:mtst}.
\begin{itemize}
    \item Short-term scheduling focuses on optimizing the daily operation of energy systems, such as the HVAC systems, lighting and plug loads.
    \item Middle-term scheduling targets energy system operations over a period, i.e., week or month, including energy storage, demand response and peak shaving. It also plays a key role in long-term energy management by addressing capacity planning, overall load estimation, and periodic energy analysis.
    \item Long-term scheduling is responsible for long-term energy-saving policy formulation and electricity procurement, aligning with the sustainability goals.
\end{itemize}
These multi-timescale strategies provide a holistic approach to managing energy systems, enhancing their adaptability and efficiency across varying temporal resolutions.
A multi-timescale optimization process is typically implemented using a rolling optimization approach, which involves solving optimization problems at different time scales iteratively.
Figure ~\ref{fig:mtso} illustrates an example of the transition from monthly planning to daily rolling planning.
The month-ahead optimization schedules overall and individual energy usage, providing periodic energy analysis. This enables building operators to analyze and optimize energy configurations and formulate medium- and long-term energy strategies.
The day-ahead optimization determines the operational schedule for the next 24 hours. The system not only considers the load information for the current day but also forecasts the load for the following day to enhance scheduling accuracy.
The intra-day rolling optimization addresses the uncertainty in energy usage. For example, Model Predictive Control (MPC) can be employed to perform rolling optimization of electrical devices throughout the day. Given the varying properties of different devices, we assume a 15-minute control period and a one-hour prediction horizon for this optimization process.

\section{Method}\label{sec:method}

\subsection{Problem Formulation and Model Overview}\label{subsec:problemsetup}

\begin{figure}[!t]
    \centering
    \includegraphics[width=\linewidth]{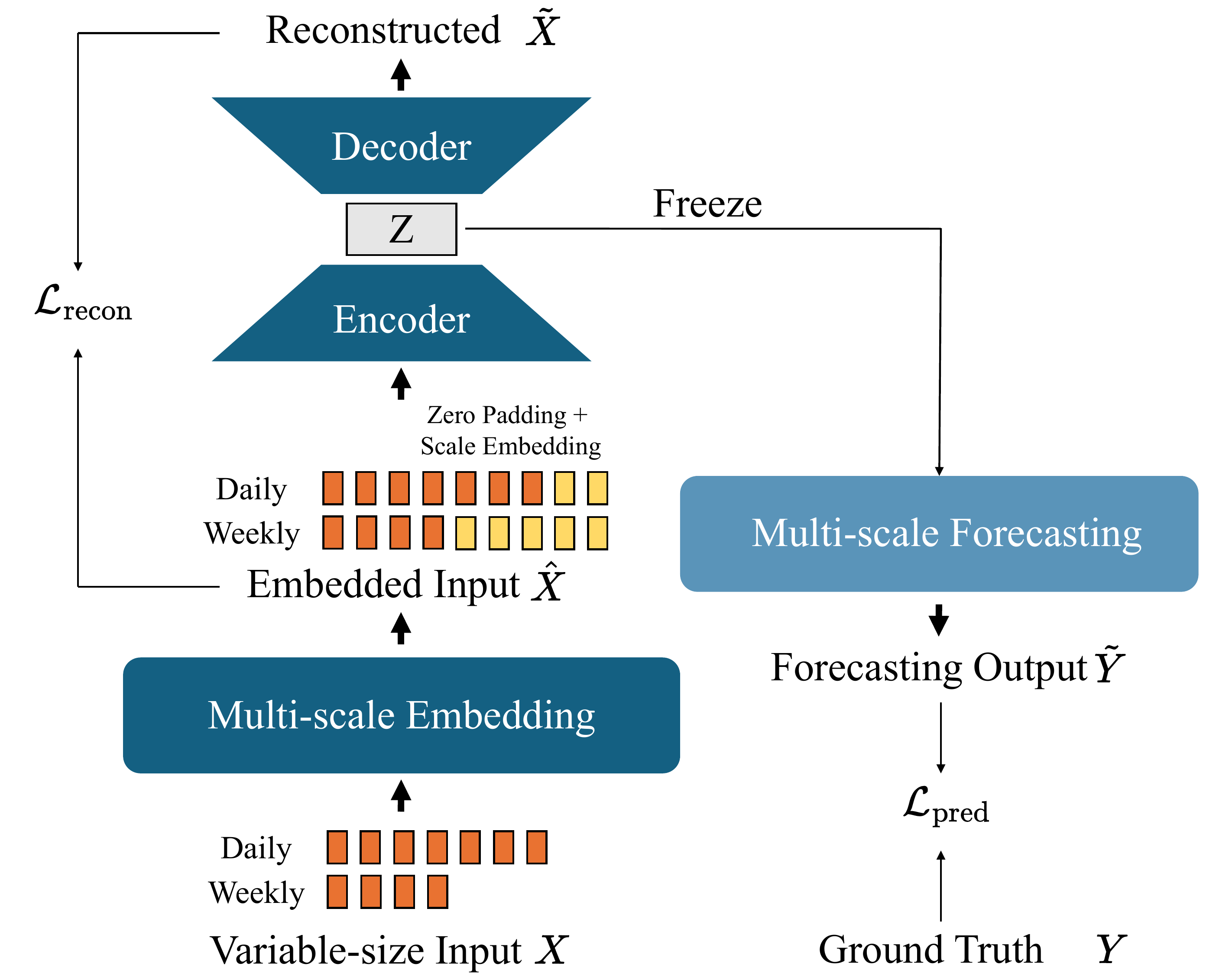}
    \caption{
    \method comprises three main components: a multi-scale embedding, a shared encoder-decoder and a multi-scale forecasting module. 
    The model is trained in two stages: initially, the encoder and decoder are jointly trained using a reconstruction loss, followed by freezing the encoder and training the forecasting module with a prediction loss.
    }
    \label{fig:overview}
\end{figure}

This section explains how \method forecasts building load at daily, weekly, monthly, and yearly levels.
Traditional prediction tasks operate on a single time scale with consistent observation intervals, represented as $Y = f(X)$, where $X \in \mathbb{R}^{L}$ denotes the input sequence of length $L$, and $Y$ denotes the target output.
In the multi-scale forecasting task, we deal with input sequences of varying lengths and need to learn representations across multiple scales. We face two main challenges:
(1) \emph{Variable Input Sizes:} due to different scales, the number of observations varies, leading to input data of different lengths. Specifically, for each scale $i \in \{1, 2, 3, 4\}$ corresponding to day, week, month, and year, the input length $L_i$ changes.
(2) \emph{Learning Representations Across Multiple Scales:} A multiscale prediction model needs to learn representations for data at different scales simultaneously,  expressed as $Y_i = f_i(X_i)$, where $X_i \in \mathbb{R}^{L_i}$ and $Y_i \in \mathbb{R}^{H}$ are the input and output at scale $i$. 
This is because the larger-scale observations aggregate those at smaller scales, indicating correlations among observations from multiple scales.
To address these challenges, we propose a method that learns a unified representation $z \in \mathbb{R}^{D}$ for data across different scales.
The overview of proposed \method as illustrated in Figure~\ref{fig:overview}.
\method comprises three main components: \text{multiscale embedding}, \text{multiscale encoder}, and \text {prediction module}.

\subsection{Multiscale Embedding}

We apply two operations to handle input data of different lengths due to varying scales: \text{zero padding} and \text{multiscale embedding}.
In \text{zero padding}, for each scale $i$, we pad the input sequence $X_i$ with zeros to reach the maximum length $L_{\text{max}} = \max_i L_i$. This ensures that all input sequences have the same length, which is important for batch processing in neural networks. The padded input is defined as:
$X_i' = [x_{i,1}, x_{i,2}, \dots, x_{i,L_i}, 0, \dots, 0] \in \mathbb{R}^{L_{\text{max}}}$,
where zeros are added to reach length $L_{\text{max}}$.

In the \text{multiscale embedding}, we create a scale embedding vector $s_i \in \mathbb{R}^I$ to represent the scale $i$. This vector helps the model distinguish between different scales. The $i$-th element of $s_i$ is 1, and the rest are zeros $s_i = [0, \dots, 1, \dots, 0]^\top$, where the $1$ is at the $i$-th position.
We then concatenate the zero-padded input $X_i'$ and the scale embedding $s_i$ to form the final embedded input for scale $i$:
\begin{equation}
\hat{X}_i = [X_i'; s_i] \in \mathbb{R}^{L_{\text{max}} + I}.
\end{equation}
This combined input includes both the time series data and the scale information, allowing the model to learn scale-specific features while maintaining a unified input format.

\subsection{Multiscale Encoder}

The shared encoder is a three-layer MLP with the ReLU activation function.
It processes the embedded inputs and obtain a unified representation $z \in \mathbb{R}^{D}$. 
We denote the input to the encoder as $\hat{X}_i$ for each scale $i$.
The encoder consists of multiple fully connected layers with activation functions defined as
$z = \text{Encoder}(\hat{X}_i)$.
The model learns a unified representation that captures common features among multiple scales by sharing the encoder across all scales. 
To ensure that $z$ effectively represents the inputs, a decoder mirroring the encoder architecture is tasked with input reconstruction. 
The decoder reconstructs the input sequences:
$\tilde{X}_i = \text{Decoder}(z)$.
We train the encoder and decoder by minimizing the reconstruction loss:
$\mathcal{L}_{\text{recon}} = \sum_{i=1}^I \| X_i' - \tilde{X}_i \|_2^2$.
This training stage helps the encoder learn a good representation $z$ that captures the essential features of the inputs. After this stage, we freeze the encoder parameters to preserve the learned representation.

\subsection{Prediction Module}

Using the frozen unified representation $z$, we train a separate prediction module to predict the target outputs $\tilde{Y} \in \mathbb{R}^{H}$. The prediction module is implemented as a fully connected layer:
\begin{equation}
\tilde{Y} = f(z) = W_{\text{pred}} z + b_{\text{pred}},
\end{equation}
where $W_{\text{pred}} \in \mathbb{R}^{H \times D}$ and $b_{\text{pred}} \in \mathbb{R}^{H}$ are the weights and biases of the prediction layer.
Our objective is to minimize the prediction loss between the predicted outputs $\tilde{Y}$ and the ground truth $Y$ as
$\mathcal{L}_{\text{pred}} = \| Y - \tilde{Y} \|_2^2$.
In many forecasting tasks, $H$ is set to 1 when predicting a single future value, such as the maximum load in the next period.
In this stage, we train only the prediction module while keeping the encoder (and decoder) parameters fixed. This approach ensures that the prediction module learns to make accurate forecasts based on the representation $z$ without altering it.

In summary, using this two-stage training procedure, we separate representation learning from the prediction task. 
The encoder learns a general representation $z$ that captures the underlying structure across different data scales. 
The prediction module then leverages this representation to make accurate forecasts.
Our method leverages the correlations among different time scales by sharing the encoder across scales and incorporating scale information through the multiscale embedding. 
This allows the model to benefit from shared patterns in the data, potentially improving forecasting accuracy.





\begin{table}[t]
\caption{MULTI-TIMESCALE POWER FORECASTING ON REAL-WORLD ELECTRICITY LOAD DATA}
\label{tab:multipofo}
\centering
\begin{tabular}{lccc}
\toprule
                             & Campus & Multi-building & Single-building \\
\midrule
Daily    &22716&3965&144016\\
Weekly   &\textbf{17083}&\textbf{3672}&93405\\
Monthly  &17551&4024&\textbf{88680}\\
\bottomrule
\end{tabular}
\end{table}
\section{Experiments}\label{sec:exp}
To evaluate  \method, we conducted extensive experiments using a real-world campus-level electricity load dataset. 
The dataset consists of electricity load measured from Suita Campus, Osaka University, Japan, spanning the years 2015 to 2020, with a temporal resolution of 30 minutes.
This dataset comprises 313 independent circuits, covering various campus facilities such as lighting, cafeterias, hospitals, libraries, and more.
We designate the initial four years (2015–2019) as the training set and the final year (2020) as the testing set.
The data was organized into four scales: daily, weekly, monthly, and yearly, each corresponding to distinct temporal resolutions. 
Min-max scaling was applied to normalize the data, ensuring consistency and enhancing model performance.
For each scale $i \in \{\text{daily}, \text{weekly}, \text{monthly}, \text{yearly}\}$, the model learns a function $f_i$ that maps an input sequence $\mathbf{X}_i$ of length $L_i$ to the predicted maximum value $\mathbf{Y}_i$:

\begin{equation}
\mathbf{X}_i = \{x_{t-L_i+1}, \dots, x_t\} \rightarrow \mathbf{Y}_i = \max(x_{t+1}, \dots, x_{t+L_i-1})
\end{equation}

To train the model, we employed a two-stage approach. In the first stage, we jointly trained the shared encoder and the reconstruction decoder using a mean squared error (MSE) loss function for 50 epochs. 
This phase ensures that the encoder effectively captures the essential features of the input data by reconstructing the original inputs. 
Once the encoder was adequately trained, we froze its parameters to retain the learned representations. 
In the second stage, we trained the prediction decoder separately for 30 epochs, optimizing it with the MSE loss to map the encoded representations to the target maximum values accurately.


\begin{figure}[!t]
    \centering
    \includegraphics[width=\linewidth]{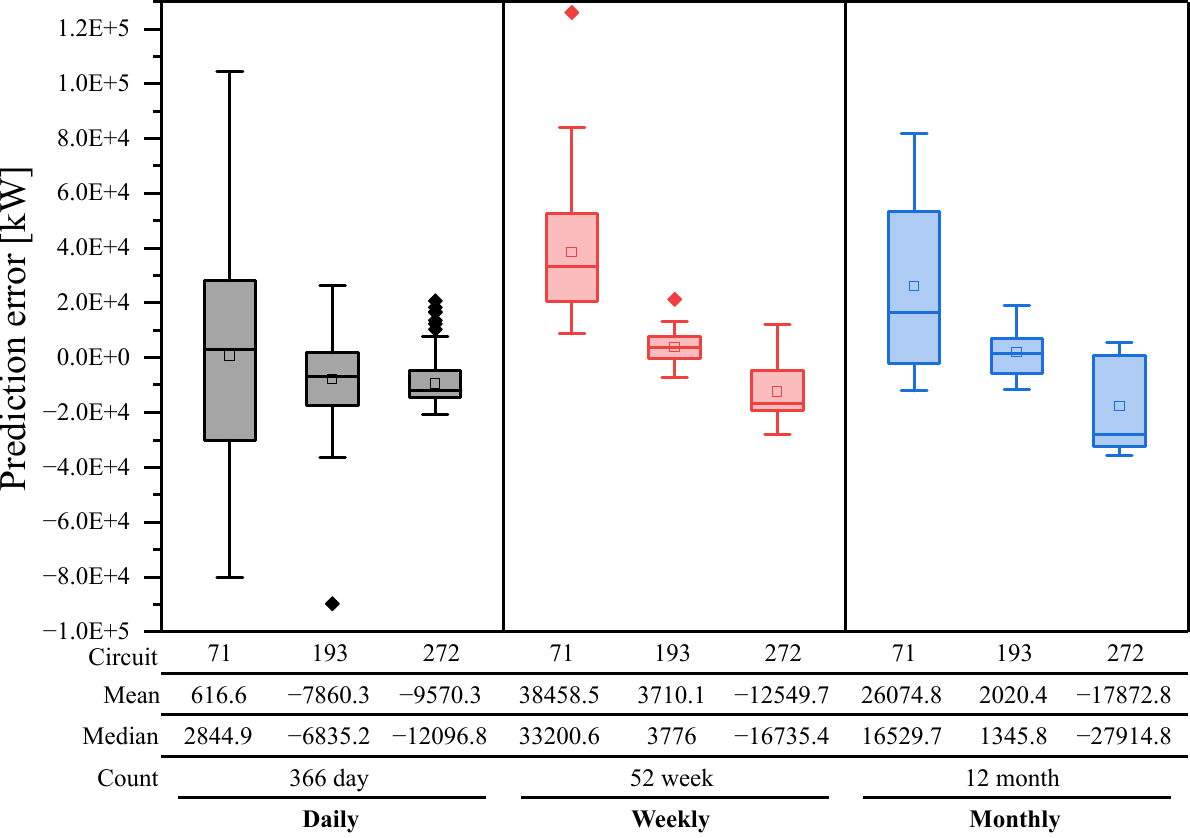}
    \caption{Prediction error distribution across daily, weekly, and monthly scales for three different circuits (71, 193, and 272). Each boxplot represents the variability in prediction error (measured in kW), highlighting the mean (square markers), median (horizontal lines within the boxes), and range (whiskers). The daily scale shows larger variability and outliers compared to weekly and monthly scales, indicating the challenge of predicting short-term energy consumption. Weekly aggregation reduces variability, while monthly aggregation further smoothens prediction errors. 
    }
    \label{fig:error-ana}
\end{figure}

Table~\ref{tab:multipofo} summarized the mean absolute error (MAE) on power forecasting at multiple scales and across multiple buildings.
The results in the table indicate that the \method provides relatively accurate power forecasts across short-, medium-, and long-term horizons. 
Since campus facilities vary widely and their electricity load characteristics differ significantly, we randomly selected three circuits from the dataset to analyze. The prediction error distribution across daily, weekly, and monthly scales is visualized in Figure~\ref{fig:error-ana}.

Figure~\ref{fig:error-ana} presents that Circuit 71 exhibits the largest variability in prediction error across different scales, while Circuits 193 and 272 demonstrate tighter distributions and relatively smaller error ranges. These results indicate that \method is capable of accurately predicting electricity load across diverse facilities with varying usage patterns.
Furthermore, the findings reveal that weekly forecasts consistently achieve the highest accuracy for both campus-wide and multi-building scenarios.
This may be because aggregating data weekly reduces noise and variability by smoothing out daily fluctuations, thereby enhancing the clarity of underlying trends and seasonal patterns. 
The weekly dependencies are pronounced due to human activity cycles, whereas monthly data primarily capture seasonal changes, and daily data carries noise and uncertainty.


\begin{figure}[!t]
    \centering
    \begin{subfigure}[b]{\linewidth}
         \centering
         \includegraphics[width=\linewidth]{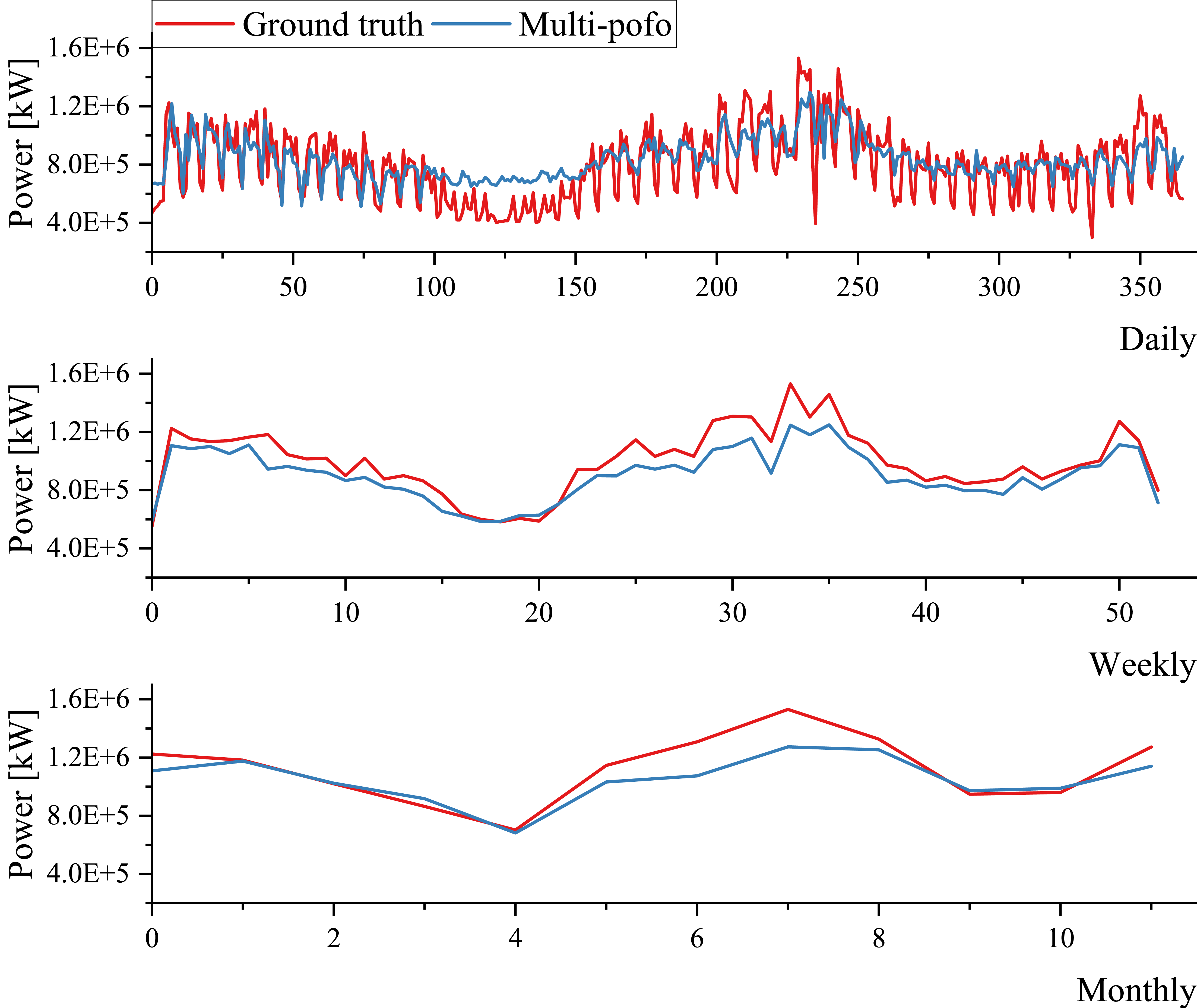}
         \caption{}
         \label{fig:pofo}
    \end{subfigure}
    \hfill
    \vspace{0.5em}
    \begin{subfigure}[b]{\linewidth}
        \centering
        \includegraphics[width=\linewidth]{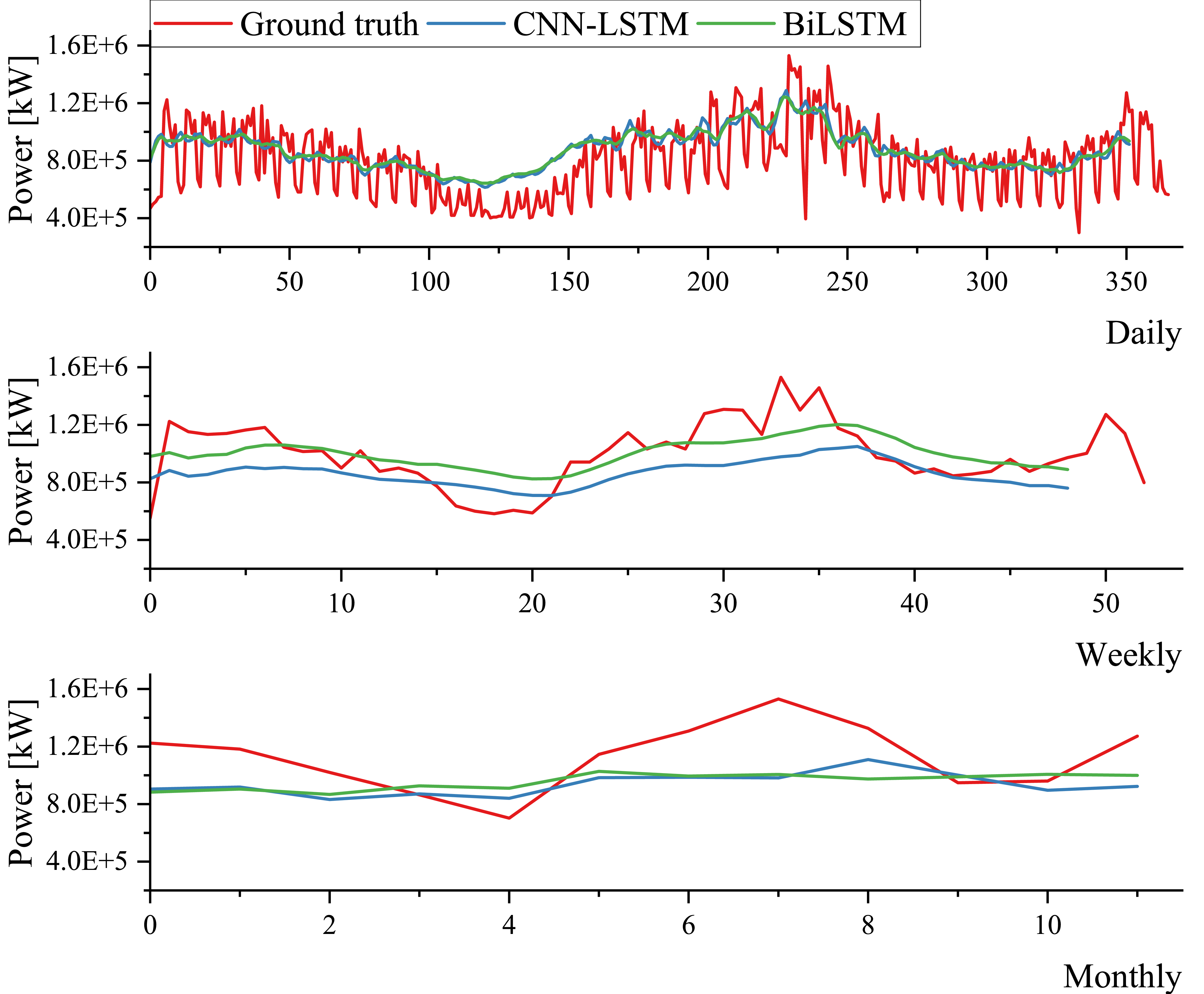}
        \caption{}
        \label{fig:cnnlstm}
    \end{subfigure}
    \caption{Comparison of load forecasting between (a) \method, and (b) CNN-LSTM, BiLSTM across day, week and month level.}
    \label{fig:results}
\end{figure}

To evaluate the performance of proposed model, we include two LSTM-base models, BiLSTM~\cite{du2020power} and CNN-LSTM~\cite{dengMultiScaleConvolutionalNeural2019} as baselines to compared with \method for single building load forecasting.
Figure~\ref{fig:results} shows the results and allow us to draw the following conclusions:
\begin{itemize}
    \item Figure~\ref{fig:pofo}  confirms the high forecasting accuracy for the single-building scenarios, as presented in Table~\ref{tab:multipofo}. The predicted weekly load curves align closely with the original load data curves.
    \item Comparing Figure~\ref{fig:pofo}
 and Figure~\ref{fig:cnnlstm}, it is evident that the LSTM struggles to accurately capture mid- to long-term temporal dependencies, particularly at the month-level. In contrast, \method effectively captures both mid- and long-term dependencies within a unified model.
\end{itemize}

\section{Conclusion}\label{sec:conclusion}
In this study, to address the challenges of multi-timescale forecasting in energy management, we proposed \method,
a unified framework for energy management that can receive varied time series inputs while providing Multi-scale power forecasting.
\method used zero padding to align the input length of different observation scales, i.e., daily, weekly, and monthly.
We also developed a novel temporal positional encoding layer to effectively index these scales. 
A simple MLP encoder then captured complex temporal dependencies across multiple timescales, enabling diverse forecasting tasks. 
Future work will focus on optimizing forecasting efficiency and conducting broader, more comprehensive evaluations.

\section*{Acknowledgment}
This work was supported by JSPS, Japan KAKENHI Grant Number 24K20901, 24K20778.

\bibliographystyle{IEEEtran}
\bibliography{ref.bib}

\end{document}